\documentclass[letterpaper,conference]{IEEEtran}
\IEEEoverridecommandlockouts

\usepackage{cite}
\usepackage{amsmath,amssymb,amsfonts}
\usepackage{graphicx}
\usepackage{textcomp}
\usepackage{xcolor}
\usepackage[nolist]{acronym}

\usepackage{graphics}
\usepackage{graphicx} 
\usepackage{multirow}
\usepackage{longtable}
\usepackage{epsfig}
\usepackage{epstopdf}

\usepackage{amsmath, mathrsfs, dsfont}
\usepackage{amssymb}
\usepackage{amsfonts}
\usepackage{bm}

\usepackage{cite}
\usepackage{verbatim}

\usepackage{amscd}
\usepackage{color}
\definecolor{lgray}{gray}{0.6}
\usepackage{rotating}

\usepackage[font=footnotesize]{subcaption}

\usepackage{mathptmx}
\usepackage[11pt]{moresize}
\usepackage{flushend}
\usepackage[stable]{footmisc}

\usepackage{algorithmicx}
\usepackage{algorithm}
\usepackage{algpascal}
\usepackage{float}
\usepackage{algc}
\usepackage{algcompatible}
\usepackage{algpseudocode}

\usepackage[vcentermath]{youngtab} 
\usepackage{url}

\newcommand{\Boldb}				{ \mathbf{b} }

\newcommand{\Boldp}				{ \mathbf{p} }

\newcommand{\Boldv}				{ \mathbf{v} }

\newcommand{\Boldx}				{ \mathbf{x} }

\newcommand{\Boldrho}			{ \boldsymbol{\rho} }

\newcommand\T{\rule{0pt}{2.6ex}}       

\newcounter{inlineenum}
\renewcommand{\theinlineenum}{\alph{inlineenum}}

\newcommand{\brmrk}[1]{\begin{remark} \label{#1} }
	\newcommand{\ermrk}{ \hfill $\bigtriangleup$    \end{remark} \vspace{1mm} }

\newtheorem{exercise}{Exercise}[section]
\newcommand{\boex}[1]{\begin{example} \label{#1} --- \rm}
	\newcommand{\eoex}{ \hfill $\bigtriangleup$    \end{example} \vspace{1mm} }
\newtheorem{example}{Example}[section]
\newcommand{\bohw}[1]{\begin{exercise} \label{#1} -- \rm}
	\newcommand{\eohw}{ \hfill    \end{exercise} \vspace{1mm} }
\newtheorem{assumption}{Assumption}[section]
\newcommand{\boass}[1]{\begin{assumption} \label{#1} -- \rm}
	\newcommand{\eoass}{ \hfill    \end{assumption} \vspace{1mm} }

\renewcommand{\baselinestretch}{1.0}

\newcommand{\black}{\color{black}}

\newcommand{\red}{\color{red}}

\definecolor{brinkpink}{rgb}{1.00, 0.33, 0.64}



\usepackage[normalem]{ulem}
\usepackage{hyperref}

\begin{document}

\title{Constrained Factor Graph Optimization for Robust Networked Pedestrian Inertial Navigation}

\author{
\IEEEauthorblockN{Yingjie Hu}
\IEEEauthorblockA{Dept. of Aerospace \\
	Engineering and Mechanics,\\
	Univ. of Minnesota, Twin Cities,\\
	Minnesota 55455, USA.\\
	Email: hu000258@umn.edu}
\and
\IEEEauthorblockN{Wang Hu}
\IEEEauthorblockA{Dept. of Electrical and\\
        Computer Engineering,\\
        Univ. of California, Riverside,\\
        CA 92521, USA.\\
	Email: whu027@ucr.edu}
}
\maketitle

\begin{acronym}
    \acro{fgo}[FGO]{Factor Graph Optimization}
    \acro{zupt}[ZUPT]{Zero-Velocity Updates}
    \acro{gnss}[GNSS]{Global Navigation Satellite Systems}
    \acro{pnt}[PNT]{Positioning, Navigation, Timing}
    \acro{imu}[IMU]{Inertial Measurement Unit}
    \acro{ins}[INS]{Inertial Navigation System}
    \acro{uwb}[UWB]{Ultra-Wideband}
    \acro{kf}[KF]{Kalman Filter}
    \acro{ekf}[EKF]{Extended Kalman Filter}
    \acro{map}[MAP]{Maximum A Posteriori}
\end{acronym}

\begin{abstract}
This paper presents a novel constrained Factor Graph Optimization (FGO)-based approach for networked inertial navigation in pedestrian localization. To effectively mitigate the drift inherent in inertial navigation solutions, we incorporate kinematic constraints directly into the nonlinear optimization framework. Specifically, we utilize equality constraints, such as Zero-Velocity Updates (ZUPTs), and inequality constraints representing the maximum allowable distance between body-mounted Inertial Measurement Units (IMUs) based on human anatomical limitations. While equality constraints are straightforwardly integrated as error factors, inequality constraints cannot be explicitly represented in standard FGO formulations. To address this, we introduce a differentiable softmax-based penalty term in the FGO cost function to enforce inequality constraints smoothly and robustly. The proposed constrained FGO approach leverages temporal correlations across multiple epochs, resulting in optimal state trajectory estimates while consistently maintaining constraint satisfaction. Experimental results confirm that our method outperforms conventional Kalman filter approaches, demonstrating its effectiveness and robustness for pedestrian navigation.
\end{abstract}


\section{Introduction}
Networked inertial navigation has recently emerged as a promising approach for pedestrian localization, leveraging the fusion of measurements from multiple \ac{imu}s. However, \ac{ins} inherently suffer from accuracy degradation over time due to the integration of measurement errors such as sensor noise, uncompensated biases, and residual biases \cite{farrell2008aided}. These accumulated errors lead to substantial position drift, particularly evident in extended navigation tasks.

A common solution to mitigate INS drift involves integrating external aiding sensors, including \ac{gnss}, \ac{uwb}, radar, and LiDAR \red \cite{groves2015principles} \black. Beyond external sensors, constraints derived from known relationships among state variables, such as kinematics, dynamics, or geometry, also provide valuable aiding information \cite{skog2010zero, yang2009kalman}. The \ac{zupt} is a prominent example of a kinematic constraint widely adopted for pedestrian navigation, exploiting the known zero velocity during the stance phase of a gait cycle \cite{skog2010zero}.

For \ac{kf}-based \ac{ins}, two principal methodologies incorporate constraints into the estimation process: the pseudo-observation approach \cite{de1997smoothly, hu2022networked} and the projection approach \cite{simon2002kalman, yang2009kalman}. The pseudo-observation method integrates equality constraints directly as additional observations, while the projection approach can handle both equality and inequality constraints by formulating and solving constrained optimization problems to yield state estimates that strictly satisfy the imposed constraints.

The \ac{fgo} approach, a nonlinear optimization framework based on \ac{map} estimation, has recently gained traction as an alternative to traditional \ac{kf}-based approaches due to their ability to comprehensively exploit temporal correlations across multiple epochs \cite{simon2006optimal, kaess2012isam2}. In \ac{fgo} frameworks, each node represents a system state at a specific epoch, with edges encoding the dynamics and measurement relationships, thus constructing an optimization problem reflecting all historical data within the considered batch. By solving this batch optimization, \ac{fgo} methods have shown robustness against cumulative drift compared to single-epoch updates typical of \ac{kf}-based methods \cite{forster2015imu, kaess2012isam2}.

As an alternative to KF-based methods, optimization-based approaches have been widely studied for PNT problems. Factor Graph Optimization (FGO) is a nonlinear optimization framework based on maximum \textit{a posteriori} (MAP) estimation \cite{simon2006optimal}. In a factor graph, each node represents the system state vector at a specific epoch, while edges encode system dynamics and measurement relationships. This leads to a nonlinear, unconstrained optimization problem that mathematically represents the factor graph. The optimal state trajectory is obtained by solving this optimization problem, the cost function of which is the sum of Mahalanobis distances (assuming Gaussian noise), representing the state transition and measurement residuals over the current data batch. Compared to conventional KF-based approaches, FGO leverages the batch of historical data and fully exploits the temporal correlations between states and observations across multiple epochs.

In this paper, we propose a constrained FGO-based approach specifically tailored for networked pedestrian inertial navigation. Our method integrates two essential types of constraints into the factor graph formulation: (1) equality constraints such as \ac{zupt}s and (2) inequality constraints representing the maximum allowable distances between body-mounted IMUs, reflective of human physiological dimensions. \ac{zupt} constraints are straightforwardly represented as error factors within the factor graph. However, inequality constraints pose challenges due to their non-direct formulation within standard FGO methods. To address this, we introduce a softmax-based penalty into the FGO cost function, providing a differentiable and stable approximation to enforce these inequality constraints effectively.

Unlike conventional constrained Kalman filters, which enforce constraints only at the current estimation epoch, our proposed constrained FGO exploits the temporal structure of data across the entire measurement batch. This approach ensures optimal trajectory estimation while consistently satisfying all constraints at every epoch, significantly mitigating position drift and enhancing localization accuracy in pedestrian navigation scenarios.

Experimental evaluations using real-world datasets demonstrate the effectiveness of our proposed constrained FGO method, showcasing considerable improvements in positioning accuracy compared to traditional constrained KF approaches. The contributions of this work are summarized as follows:

\begin{itemize}
\item Development of a novel FGO-based networked inertial navigation framework integrating both equality and inequality constraints.
\item Proposal of a softmax-based penalty method within the FGO framework for incorporating inequality constraints related to human biomechanical limits.
\item Experimental validation demonstrating superior navigation accuracy and robustness over conventional KF-based methods.
\end{itemize}

\section{Problem Statement}
The personal navigation problem considered in this work involves fusing measurements from body-mounted inertial sensors to estimate a pedestrian’s trajectory. The proposed networked pedestrian navigation system utilizes two foot-mounted IMUs as the primary navigation sensors. To improve estimation accuracy and reduce drift, the sensor fusion framework incorporates both \ac{zupt} constraints and a constraint on the maximum distance between the feet.

In multi-agent \ac{pnt} systems, two common data fusion architectures are centralized and decentralized. In a decentralized architecture, each agent runs an independent estimator, and inter-agent correlations are indirectly established through occasional information exchange. However, decentralized systems generally lack mechanisms to explicitly manage cross-agent correlations. In contrast, a centralized architecture maintains a single global estimator that jointly models all agent states, allowing inter-agent correlations to be preserved directly through the global covariance matrix. This centralized structure enables all agents to benefit collectively from shared measurements and constraints. Therefore, a centralized fusion scheme is adopted for the two-IMU configuration in this study.

The navigation system consists of two IMU nodes. The state vector for each individual IMU is defined as:
\begin{align} \label{eq:01}
    {}^{s}\Boldx_i = \left[{}^{s}\Boldp_i;\, {}^{s}\Boldv_i;\, {}^{s}\Boldrho_i;\, {}^{s}\Boldb_{a,i};\, {}^{s}\Boldb_{g,i} \right]
\end{align}
where the superscript $s$ refers to the $s$-th IMU ($s = 1,2$). Each state vector is $16 \times 1$ and includes the position $\Boldp$, velocity $\Boldv$, quaternion $\Boldrho$, and accelerometer biases $\Boldb_{a}$, and gyroscope biases $\Boldb_{g}$. The full state vector of the centralized system is formed by concatenating the individual IMU state vectors:
\begin{align}\label{eq:02}
    \Boldx = [{}^1\Boldx_i;\, {}^2\Boldx_i].
\end{align}
State propagation for each IMU is performed using standard inertial navigation mechanization. Relevant references for \ac{INS} mechanization can be found in \cite{groves2015principles}.

GNSS signals of good quality are not always available in personal navigation scenarios due to environmental limitations, such as indoor settings or deep urban canyons. In such cases, other external aiding sensors, e.g., cameras and UWB, can be utilized to help the navigation. In addition to external aiding sensors, constraints derived from kinematics/dynamics can be employed to assist navigation. Zero-velocity update (ZUPT) is one of the most commonly used kinematic constraints in personal navigation and automotive navigation. In personal navigation, a subject's gait cycle alternates between the stance phases and the swing phases. During the swing phase, the corresponding foot is above the ground and moving in the air until it touches the ground. In comparison, the stance phase refers to the period in which the foot is in contact with the ground. An assumption can be made that the instantaneous velocity of the food in the stance phase is zero. This zero-velocity information serves as a pseudo-measurement that can be incorporated into the estimation framework to reduce position drift. Because the gait cycle alternates between these phases, ZUPT measurements become periodically available throughout the walking process.

ZUPT constraints are incorporated into the filtering framework as pseudo-measurements. Another source of constraints comes from the physical limitation on the size of the human body. For instance, the maximum foot separation of a human subject has an upper bound. For the two foot-mounted IMUs, this inter-foot distance constraint can be applied to limit the relative positions of the IMU nodes and potentially reduce drift. Unlike ZUPT, which can be treated as a direct pseudo-measurement, the inter-foot separation constraint takes the form of an inequality constraint. To incorporate inequality constraints into the filtering process, the projection approach is commonly used \cite{skog2012fusing}. This method projects the unconstrained state estimate onto the constraint surface to satisfy the constraint. The constraint Kalman filter can be usually formulated as the constrained optimization problem as follows:
\begin{equation}
    \begin{aligned} \label{eq:03}
        \hat{\mathbf{x}}^{*} = \arg\min_{\mathbf{x}} (\mathbf{x} - \hat{\mathbf{x}}^{+})^{T}\left({\mathbf{P}^{+}}\right)^{-1}(\mathbf{x} - \hat{\mathbf{x}}^{+}) \quad \text{s.t.} \quad \mathbf{D}(\mathbf{x}) \leq \mathbf{d}
    \end{aligned}
\end{equation}
in which, $\hat{\mathbf{x}}^{+}$ denotes the unconstrained state estimate and its corresponding error covariance matrix is $\mathbf{P}^{+}$. The state has to satisfy the inequality constraint $\mathbf{D}(\mathbf{x}) \leq \mathbf{d}$. Solving the constrained optimization problem in \eqref{eq:03} yields the constrained state estimate $\hat{\mathbf{x}}^{*}$.

\section{Methodology}
In this paper, we consider a dual-IMU networked inertial navigation system, configured by mounting one IMU on each foot of the pedestrian. As a result, ZUPT constraints are available independently for both IMUs. When a foot is in full contact with the ground, the instantaneous velocity of the corresponding IMU is assumed to be zero. The zero-velocity detection is performed based on the magnitude of IMU measurements from accelerometers and gyroscopes. Specifically, epochs for applying ZUPT constraints are identified whenever the norm of these IMU sensor outputs drops below predefined thresholds. Additionally, the inter-foot distance serves as another kinematic constraint. Given the physical limitations imposed by human limb length, an upper bound naturally exists on the distance between the two foot-mounted IMUs. This inter-foot distance constraint is modeled as an inequality constraint on the positions of the IMU nodes. Unlike ZUPT constraints, which are dynamically activated based on detected foot stance phases, the inter-foot distance constraint is enforced periodically at fixed intervals to ensure consistent adherence to the upper bound throughout the entire navigation period.

\begin{figure}[bt]
\centerline{\includegraphics[width=9cm]{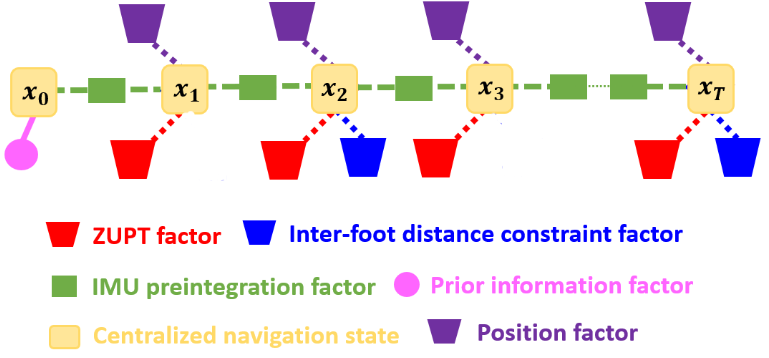}}
\caption{Architecture of the proposed FGO-based personal navigation algorithm. The subscript $1:T$ denotes the indices within the sliding window.}
\label{fig:fgo}
\end{figure}

A factor graph is constructed to model the dual-IMU navigation system for pedestrian localization, incorporating these kinematic constraints to mitigate position drift. The architecture of the proposed FGO-based personal navigation algorithm is illustrated in Fig. \ref{fig:fgo}. It should be noted that not all epochs include position, ZUPT, and inter-foot distance constraint factors simultaneously. The presence of each factor type depends on data availability and scheduling: position factors are integrated only when external position measurements are available; ZUPT factors are dynamically activated upon detecting zero-velocity conditions; and inter-foot distance constraint factors are applied at predetermined fixed intervals. The centralized navigation state is defined in \eqref{eq:02}, containing the $16 \times 1$ state vector for each IMU. The remainder of this section will elaborate on each factor involved in the FGO architecture to formulate the nonlinear optimization clearly.

\textbf{IMU Preintegration Factor}: The INS mechanization equations govern how each IMU's navigation states propagate over time. These equations impose dynamic constraints on the transition between consecutive states. However, due to the high sampling rate of IMU measurements, the computational load of FGO will become prohibitive if all the IMU states are to be processed. To alleviate the computational burden, a technique called IMU preintegration is introduced \cite{forster2015imu}, which allows the integration of high-frequency IMU data without requiring every measurement to be directly included in the optimization problem, significantly reducing computational complexity. The resulting IMU preintegration factor is denoted by $\| e_{\text{\tiny IMU},i} \|^2$ and the detailed derivation of $\| e_{\text{\tiny IMU},i} \|^2$ can be found in \cite{forster2015imu},\cite{forster2016manifold}.

\textbf{Zero Velocity Factor}: ZUPT constraints are represented as zero velocity factors at epochs when a foot is in contact with the ground, implying zero instantaneous velocity. The zero-velocity factor encodes the velocity information for the corresponding IMU states.
\begin{equation}
    \begin{aligned} \label{eq:04}
        \| e_{\text{\tiny ZUPT}, i} \|^2_{\mathbf{R}_{\text{\tiny ZUPT},i}} & = \| \mathbf{y}_{\text{\tiny ZUPT},i} - \mathbf{H}_{\text{\tiny ZUPT}}\mathbf{x}_{i} \|^2_{\mathbf{R}_{\text{\tiny ZUPT},i}}. \\
    \end{aligned}
\end{equation}
where $\mathbf{y}_{\text{\tiny ZUPT},i}$ represents the zero velocity measurement, that is, $\mathbf{y}_{\text{\tiny ZUPT},i} = \mathbf{0}$. $\mathbf{R}_{\text{\tiny ZUPT},i}$ denotes the covariance matrix of the zero velocity measurement noise. The measurement matrix $\mathbf{H}_{\text{\tiny ZUPT}}$ can be determined as:
\begin{itemize}
    \item When the right foot is in stance phase:
    \begin{equation}
        \begin{aligned} \label{eq:H1}
            \mathbf{H}_{\text{\tiny ZUPT}} = \begin{bmatrix}
                \mathbf{0}_{3 \times 3} & \mathbf{I}_{3 \times 3} & \mathbf{0}_{3 \times 9} & \mathbf{0}_{3 \times 15} \\
            \end{bmatrix}
        \end{aligned}
    \end{equation}
    \item When left foot is in stance phase:
    \begin{equation}
        \begin{aligned} \label{eq:H2}
            \mathbf{H}_{\text{\tiny ZUPT}} = \begin{bmatrix}
                \mathbf{0}_{3 \times 15} & \mathbf{0}_{3 \times 3} & \mathbf{I}_{3 \times 3} & \mathbf{0}_{3 \times 9} \\
            \end{bmatrix}
        \end{aligned}
    \end{equation}
    \item When both right and left feet are in stance phase:
    \begin{equation}
        \begin{aligned}
            \mathbf{H}_{\text{\tiny ZUPT}} = \begin{bmatrix}
                \mathbf{0}_{3 \times 3} & \mathbf{I}_{3 \times 3} & \mathbf{0}_{3 \times 9} & \mathbf{0}_{3 \times 15} \\
                \mathbf{0}_{3 \times 15} & \mathbf{0}_{3 \times 3} & \mathbf{I}_{3 \times 3} & \mathbf{0}_{3 \times 9} \\
            \end{bmatrix}
        \end{aligned}
    \end{equation}
\end{itemize}

\textbf{Inter-Foot Distance Constraint Factor}: Due to the limitation on the physical size of the human body, the distance between two feet cannot be greater than a certain upper bound. This inter-foot distance constraint can be expressed by
\begin{equation}
    \begin{aligned} \label{eq:05}
        \| {}^{1}\mathbf{p}_{i} - {}^{2}\mathbf{p}_{i} \|_{2} - d \leq 0
    \end{aligned}
\end{equation}
where, ${}^{1}\mathbf{p}_{i}$ and ${}^{2}\mathbf{p}_{i}$ denote the position vectors of the two foot-mounted IMUs, $\| \cdot \|_2$ is the 2-norm operator and $d$ represents the upper bound of the inter-foot distance. Unlike the zero velocity constraint, equation~\eqref{eq:05} is an explicit inequality constraint, which cannot be directly incorporated into the FGO formulation as constraint factors. To address this issue, we propose augmenting the nonlinear graph optimization with the softmax penalty that represents the inequality distance constraint. The softmax penalty of the distance constraint can be expressed as
\begin{subequations}
    \begin{align}
    \label{eq:softmax1}
        \text{softmax} \left( \Delta d_{i} \right) & =  \frac{1}{\alpha} \text{log} \left( 1 + e^{\alpha \Delta d_{i}} \right) \\ 
        \Delta d_{i} &= \| {}^{1}\mathbf{p}_{i} - {}^{2}\mathbf{p}_{i} \|_{2} - d
    \end{align}
\end{subequations}
where, parameter $\alpha$ is the sharpness variable that controls how closely the softmax function approximates the hard max function max$\left( 0,\Delta d_{i} \right)$. The hard max function aligns with the active set method \cite{kim2008nonnegative} for constrained optimization problems. When $\Delta d = \| {}^{1}\mathbf{p}_{i} - {}^{2}\mathbf{p}_{i}\|_{2} - d  > 0$, the constraint \eqref{eq:05} is not satisfied and is, thus, an active constraint. Accordingly, the hard max penalty max$\left( 0, \Delta d_{i} \right) = \Delta d_{i}$ is augmented to penalize the cost function. On the other hand, when $\Delta d_{i} = \| {}^{1}\mathbf{p}_{i} - {}^{2}\mathbf{p}_{i} \|_{2} - d  \leq 0$, the constraint \eqref{eq:05} is met and becomes an inactive constraint. The hard max penalty reduces to zero, indicating that the constraint has no impact on the optimization. However, the hard max penalty is not differentiable at $\Delta d_{i} = 0$, which can cause instability in gradient-based optimization when solving the FGO. In comparison, the softmax penalty approach is smooth and differentiable, and more numerically stable, making it better suited for FGO solving using gradient-based solvers, such as Levenberg-Marquardt, Gauss-Newton. 

\textbf{Position Factor:} In addition to the ZUPT and inter-foot distance constraint, position measurements are formulated as factors when they become available. The position factor is represented by
\begin{equation}
    \begin{aligned} \label{eq:gnss}
        \| e_{\text{\tiny pos},i} \|^2_{\mathbf{R}_{\text{pos},i}} = \| \mathbf{y}_{\text{\tiny pos},i} - \mathbf{H}_{\text{\tiny pos}}\mathbf{x}_{i} \|^2_{\mathbf{R}_{\text{\tiny pos},i}}
    \end{aligned}
\end{equation}
The loosely coupled GNSS/INS integration is adopted in this paper where the IMU and position information are fused on the navigation solution level. $\mathbf{y}_{\text{\tiny pos},i}$ denotes the position measurements and ${\mathbf{R}_{\text{\tiny pos},i}}$ is the covariance matrix for the position measurement noise. The measurement matrix $\mathbf{H}_{\text{\tiny pos}}$ is
\begin{equation}
    \begin{aligned} 
        \mathbf{H}_{\text{\tiny pos}} = \begin{bmatrix}
            \bf{I}_{3 \times 3} & \bf{0}_{3 \times 12} & \bf{0}_{3 \times 3} & \bf{0}_{3 \times 12} \\
            \bf{0}_{3 \times 3} & \bf{0}_{3 \times 12} & \bf{I}_{3 \times 3} & \bf{0}_{3 \times 12} \\
        \end{bmatrix}
    \end{aligned}
\end{equation}

\textbf{Prior Information Factor:} The prior information can be formulated as a factor incorporated into the FGO formulation. For instance, if the $\textit{a priori}$ state estimate at the first time step is $\hat{\mathbf{x}}^{-}_{0}$ and the corresponding error covariance matrix is $\mathbf{P}^{-}_{0}$, the prior information factor can be written as
\begin{equation}
    \begin{aligned} \label{eq:pri}
        \| e_{\text{pri}} \|^2_{\mathbf{P}^{-}_{0}} = \| \mathbf{x}_0 - \hat{\mathbf{x}}^{-}_{0} \|^2_{\mathbf{P}^{-}_{0}}.
    \end{aligned}
\end{equation}

Using \eqref{eq:04}, \eqref{eq:softmax1}, \eqref{eq:gnss}, and \eqref{eq:pri}, the cost function of the FGO can be formulated as
\begin{equation}
    \begin{aligned}
        \hat{\mathbf{x}}_{\text{0:T}} = \operatorname*{argmin}_{  \mathbf{x}_{\text{0:T}}} & \| e_{\text{pri}} \|^2_{\mathbf{P}^{-}_{0}}  + \sum_{i=1}^{T}( a_{i} \| e_{\text{\tiny ZUPT}, i} \|^2_{\mathbf{R}_{\text{\tiny ZUPT},i}} + b_{i} \| e_{\text{\tiny pos},i} \|^2_{\mathbf{R}_{\text{pos},i}}) \\ 
        & + \sum_{i=1}^{T-1} \| e_{\text{\tiny IMU},i} \|^2_{\mathbf{Q}_{\text{\tiny IMU},i}} + \sum_{i=1}^{T} \lambda_{i} \text{softmax}\left( 
\Delta d_{i} \right) . \\
    \end{aligned}
\end{equation}
in which, $a_i \in \{0,1\}$ and $b_i \in \{0,1\}$ are the parameters indicating if ZUPT and position measurements are available for the current time step $i$. For example, $a_i = 1$ indicates that ZUPT measurement is available and $b_i = 0$ indicates that there are no position measurements for the current time step $i$. $\lambda_{i}$ is the penalty weight parameter for the softmax penalty at time step $i$. The proposed FGO-based personal navigation algorithm is implemented using the open-source GTSAM \cite{dellaert2012factor} package in Python. In order to increase the computational efficiency of the proposed algorithm, incremental nonlinear optimization 2 algorithms (iSAM2) \cite{kaess2012isam2} is used in our implementation to solve the nonlinear graphical optimization.
\section{Experiment}
To evaluate the effectiveness of the proposed constrained \ac{fgo}-based algorithm for pedestrian localization, a real-world experiment was conducted. The proposed method was compared with an \ac{ekf} approach that incorporates position measurements, \ac{zupt}, and inter-foot distance constraints. In the \ac{ekf} implementation, \ac{zupt} is integrated using the pseudo-observation method, while the inter-foot distance constraint is handled via the projection approach. Position measurements, when available, are fused into the \ac{ekf} using a loosely coupled integration scheme.

\begin{figure}[htbp]
\centerline{\includegraphics[width=\linewidth]{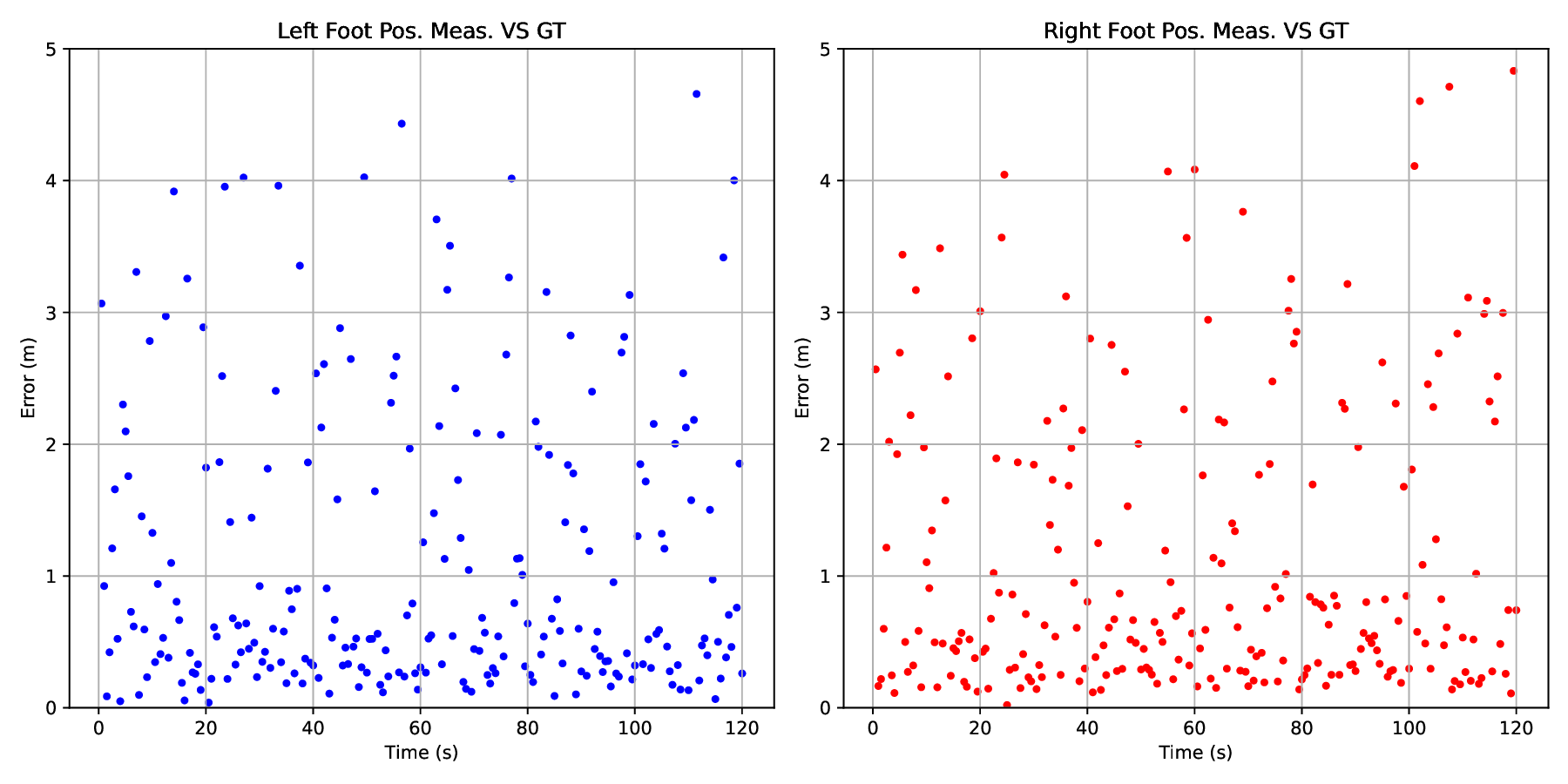}}
\caption{Position measurements horizontal errors.}
\label{fig:position_meas_errors}
\end{figure}

Data collection was carried out in the Shepherd Drone Lab at the University of Minnesota, which is equipped with a high-precision optical motion capture system (PhaseSpace Impulse X2E)~\cite{PhaseSpace2025MoCap}. This system tracks the positions of LED markers with centimeter-level of accuracy, and was used to obtain ground truth trajectories for the evaluation. To simulate real-world conditions, noisy and corrupted position measurements were generated from the ground-truth trajectory. Specifically, zero-mean Gaussian noise with a standard deviation of 30 centimeters was added to the true positions, mimicking typical measurement noise in real-world GNSS data. Additionally, outliers were introduced by injecting occasional large biases randomly selected from a predefined set of values. These outliers emulate the severe errors that can occur in GNSS-denied or multipath-rich environments \cite{wang2024rapsrtk}, such as urban canyons or indoor settings. For the purpose of this study, position measurements were assumed to be available at 0.5-second intervals\footnote{The source code for the experiment is available at \url{https://github.com/Azurehappen/Constrained-ZUPT-FGO}}. The horizontal-plane errors of these corrupted position measurements for both feet are shown in Fig.~\ref{fig:position_meas_errors}.

Xsens DOT units~\cite{xsens2025DoT} were used as the \ac{imu}s in this experiment, capable of providing raw specific force and angular velocity data. One \ac{imu} was mounted on each foot of the subject. LED markers were affixed on top of the \ac{imu}s to track their global positions within the motion capture system's reference frame. The \ac{imu}s recorded data at 60 Hz, while the motion capture system operated at 960 Hz. Post-processing was required to synchronize the two data streams.

The human subject wearing the \ac{imu}s walked throughout the test area, and all collected data were processed offline. The same dataset was used to evaluate both the proposed ac{fgo}-based method and the \ac{ekf} approach. The following methods were compared:

\begin{itemize}
    \item \textsc{ekf-zupt}: EKF with ZUPT constraints;
    \item \textsc{ekf-zupt-step}: EKF with ZUPT and inter-foot distance constraints;
    \item \textsc{ekf-zupt-pos}: EKF with ZUPT and position measurements;
    \item \textsc{ekf-zupt-pos-step}: EKF with ZUPT, inter-foot distance, and position measurements;
    \item \textsc{fgo-zupt}: FGO with ZUPT constraints;
    \item \textsc{fgo-zupt-step}: FGO with ZUPT and inter-foot distance constraints;
    \item \textsc{fgo-zupt-pos}: FGO with ZUPT and position measurements;
    \item \textsc{fgo-zupt-pos-step}: FGO with ZUPT, inter-foot distance, and position measurements;
\end{itemize}

\begin{figure}[htbp]
\centerline{\includegraphics[width=\linewidth]{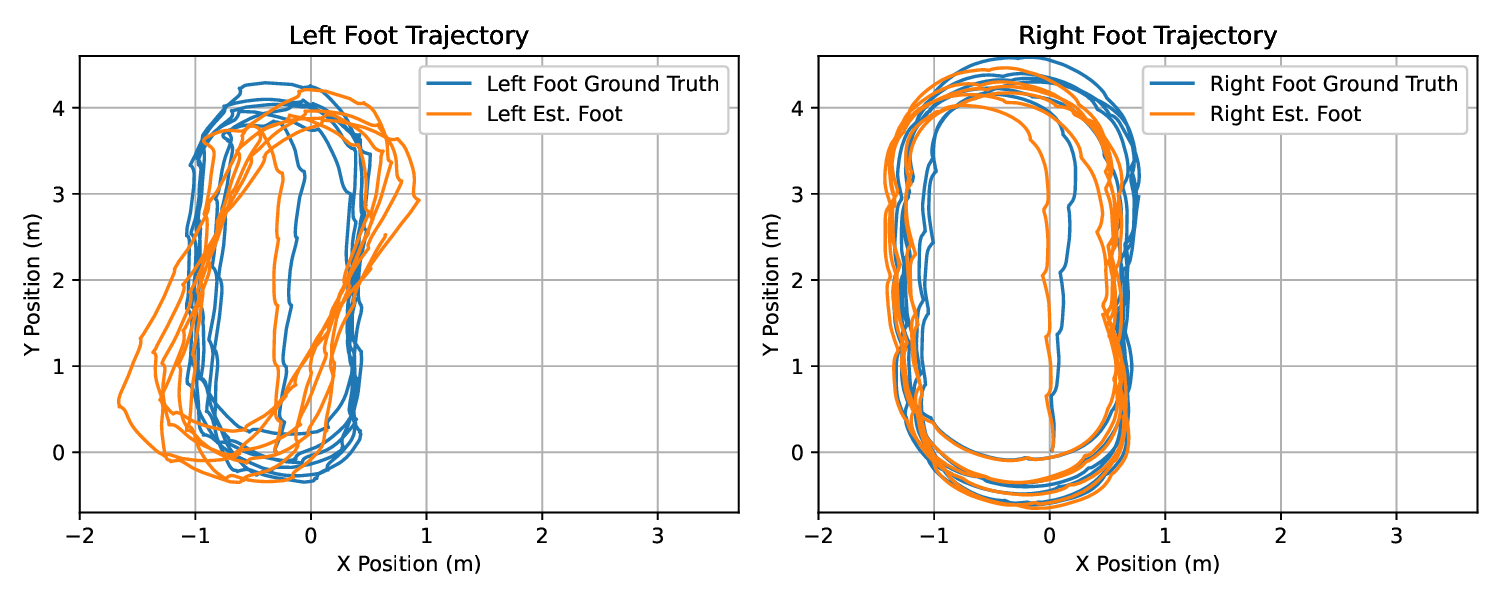}}
\caption{The comparison between the ground truth (blue) and the estimated trajectory of \textsc{FGO-ZUPT-POS-STEP} (orange).}
\label{fig:fgo_traj_errors}
\end{figure}

\begin{figure}[htbp]
\centerline{\includegraphics[width=\linewidth]{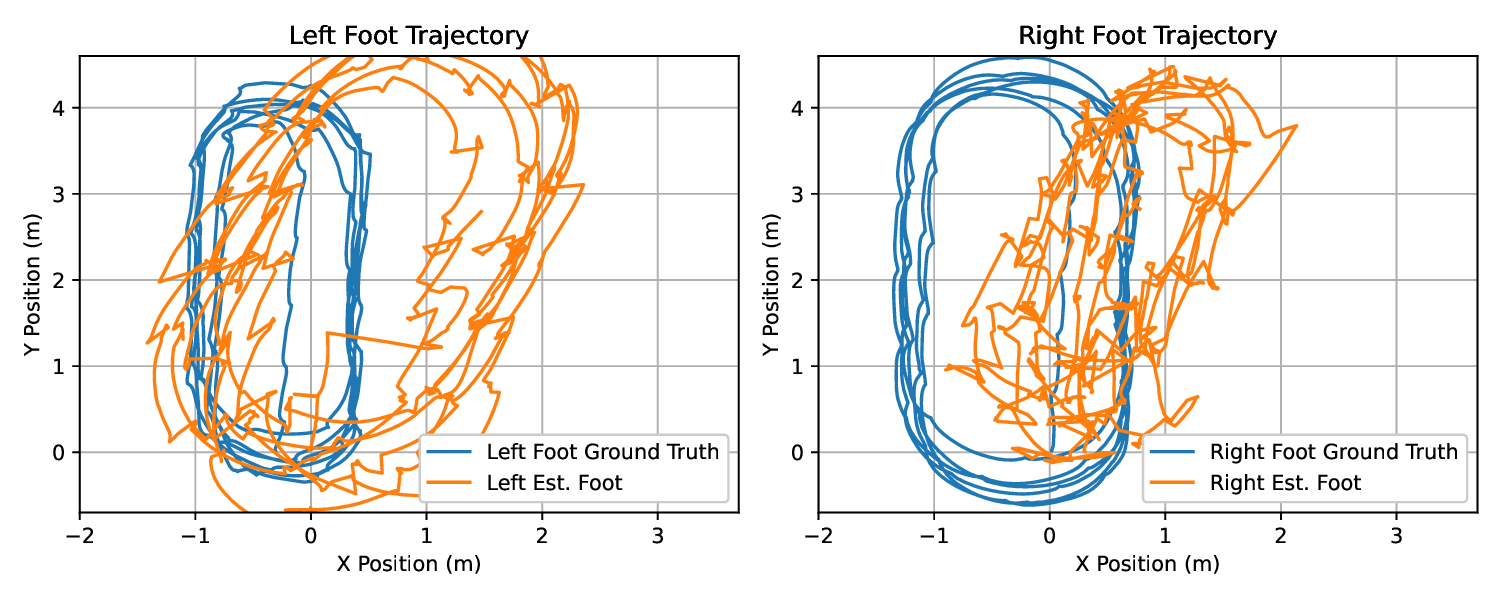}}
\caption{The comparison between the ground truth (blue) and the estimated trajectory of \textsc{EKF-ZUPT-POS-STEP} (orange).}
\label{fig:ekf_traj_errors}
\end{figure}

Fig.~\ref{fig:fgo_traj_errors} shows the estimated trajectory of \textsc{FGO-ZUPT-POS-STEP} (orange) compared to the ground truth trajectory (blue) for a 2-minute walking session. The estimated trajectory follows the ground truth with good accuracy. In comparison, Fig.~\ref{fig:ekf_traj_errors} illustrates the estimated trajectory using \textsc{EKF-ZUPT-POS-STEP} for the same dataset, which exhibits relative larger position errors. Performance comparisons among all methods are summarized in Tables~\ref{tab:horr_right} through~\ref{tab:horper_left}, using metrics including mean error, root-mean-square (RMS) error, and maximum error. Among all evaluated methods, \textsc{FGO-ZUPT-POS} achieves the lowest errors across all three metrics.

\begin{figure}[htbp]
\centerline{\includegraphics[width=\linewidth]{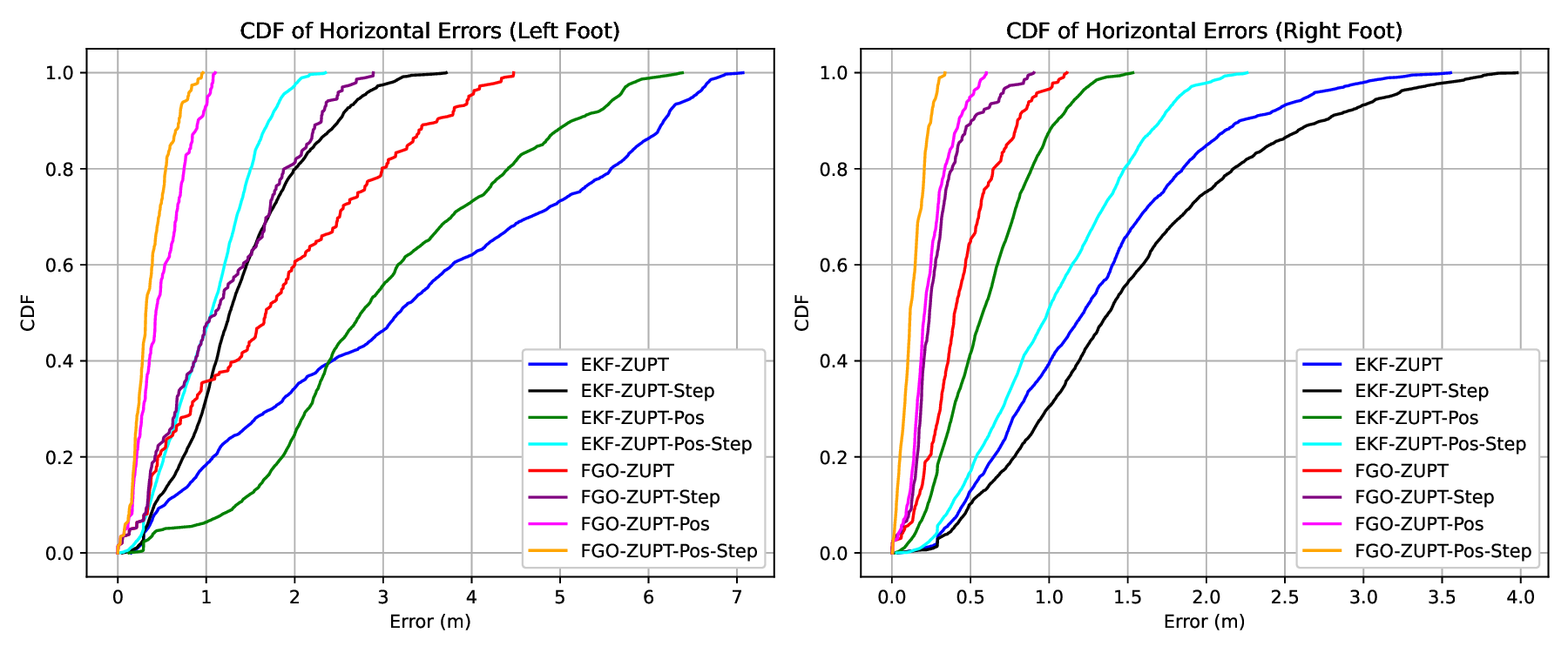}}
\caption{The cumulative distribution function of horizontal position errors for EKF-based and FGO-based approaches for each foot.}
\label{fig:cdf_errors}
\end{figure}

Additionally, the cumulative distribution functions (CDFs) of the horizontal position errors are plotted in Fig.~\ref{fig:cdf_errors} for both \ac{ekf}- and \ac{fgo}-based methods. The rightmost CDF curve corresponds to the proposed \textsc{FGO-ZUPT-POS}, demonstrating the smallest horizontal error distribution among all approaches. These experimental results validate the effectiveness of the proposed constrained \ac{fgo}-based algorithm and show that it achieves superior positioning accuracy compared to traditional constrained Kalman filter methods.

\begin{table}[htbp]
    \centering
    \begin{tabular}{lccc}
        \hline
        \textbf{Method}\T & \textbf{Mean (m)} & \textbf{RMS (m)} & \textbf{Max (m)} \\
        \hline
        EKF-ZUPT \T     & 1.310&	1.484&	3.554 \\
        EKF-ZUPT-STEP  \T     & 1.525  & 1.730  & 3.977  \\
        EKF-ZUPT-POS \T  & 0.631  & 0.706  & 1.533  \\
        EKF-ZUPT-POS-STEP \T  & 1.029  & 1.134  & 2.259  \\
        FGO-ZUPT  \T     & 0.446  & 0.510  & 1.113  \\
        FGO-ZUPT-STEP\T   & 0.282  & 0.331  & 0.903  \\
        FGO-ZUPT-POS  \T     & 0.238  & 0.270  & 0.601  \\
        FGO-ZUPT-POS-STEP\T   & \textbf{0.131}  & \textbf{0.153}  & \textbf{0.337}  \\
        \hline
    \end{tabular}
    \caption{Horizontal Error Statistics for the Right Foot.}
    \label{tab:horr_right}
\end{table}

\begin{table}[htbp]
    \centering
    \begin{tabular}{lccc}
        \hline
        \textbf{Method}\T & \textbf{Mean (m)} & \textbf{RMS (m)} & \textbf{Max (m)} \\
        \hline
        EKF-ZUPT   \T    & 3.338&	3.904&	7.085 \\
        EKF-ZUPT-STEP   \T    & 1.427&	1.607&	3.713  \\
        EKF-ZUPT-POS  \T & 3.057 &	3.367 &	6.414  \\
        EKF-ZUPT-POS-STEP  \T & 1.068 &	1.181 &	2.348  \\
        FGO-ZUPT \T      & 1.788 &	2.169 &	4.474 \\
        FGO-ZUPT-STEP \T  & 1.201 &	1.414 &	2.888  \\
        FGO-ZUPT-POS \T  & 0.488&	0.566&	1.101
  \\
        FGO-ZUPT-POS-STEP \T  & \textbf{0.368}  & \textbf{0.425}  & \textbf{0.963}  \\
        \hline
    \end{tabular}
    \caption{Horizontal Error Statistics for the Left Foot.}
    \label{tab:horr_left}
\end{table}

\begin{table}[htbp]
    \centering
    \begin{tabular}{lccc}
        \hline
        \textbf{Method}\T & \textbf{$90\%$ (m)} & \textbf{$95\%$ (m)} & \textbf{$99\%$ (m)} \\
        \hline
        EKF-ZUPT  \T   &  2.269 &	2.656 &	3.235
  \\
        EKF-ZUPT-STEP  \T     & 2.723 &	3.115 &	3.694
  \\
        EKF-ZUPT-POS \T  & 1.078 &	1.206 &	1.363
  \\
        EKF-ZUPT-POS-STEP \T  & 1.699 &	1.827 &	2.107
  \\
        FGO-ZUPT  \T     & 0.799 &	0.904 &	1.088
  \\
        FGO-ZUPT-STEP\T   & 0.507 &	0.688 &	0.862
  \\
        FGO-ZUPT-POS  \T     & 0.427 &	0.499 &	0.585
  \\
        FGO-ZUPT-POS-STEP\T   & \textbf{0.248}&	\textbf{0.281}&	\textbf{0.305}
  \\
        \hline
    \end{tabular}
    \caption{$90\%$, $95\%$ and $99\%$ error bound for the Right Foot.}
    \label{tab:horper_right}
\end{table}

\begin{table}[htbp]
    \centering
    \begin{tabular}{lccc}
        \hline
        \textbf{Method}\T & \textbf{$90\%$ (m)} & \textbf{$95\%$ (m)} & \textbf{$99\%$ (m)} \\
        \hline
        EKF-ZUPT   \T    & 6.177 &	6.483 &	6.771
  \\
        EKF-ZUPT-STEP   \T    & 2.515 &	2.777 &	3.223
  \\
        EKF-ZUPT-POS  \T & 5.111 &	5.597 &	5.998
  \\
        EKF-ZUPT-POS-STEP  \T & 1.744 &	1.890 &	2.133
  \\
        FGO-ZUPT \T      & 3.609&	3.968&	4.362
  \\
        FGO-ZUPT-STEP \T  & 2.300 &	2.479 &	2.809
  \\
        FGO-ZUPT-POS \T  & 0.904 &	1.013 &	1.077
  \\
        FGO-ZUPT-POS-STEP \T  & \textbf{0.692}&	\textbf{0.793}&	\textbf{0.944}
  \\
        \hline
    \end{tabular}
    \caption{$90\%$, $95\%$ and $99\%$ error bound for the Left Foot.}
    \label{tab:horper_left}
\end{table}



\section{Conclusion}
This paper presented a constrained \ac{fgo}-based networked inertial navigation system designed for pedestrian localization. Unlike conventional Kalman Filter-based methods, the proposed approach fully exploits temporal correlations across states, observations, and constraints throughout the entire batch of navigation data. Specifically, we introduced a novel integration strategy for inequality constraints by augmenting the nonlinear optimization problem with a differentiable softmax penalty term, ensuring stable and efficient optimization.

Experimental validation conducted using real-world pedestrian datasets demonstrated the superior performance of our proposed method. Quantitative results clearly show that the constrained FGO-based algorithm consistently outperforms conventional constrained Kalman filter approaches, yielding significantly lower mean, RMS, and maximum positioning errors. These improvements highlight the benefits of explicitly modeling and exploiting temporal and constraint correlations across multiple epochs within a unified optimization framework.

\bibliographystyle{IEEEtran.bst}
\bibliography{IEEEabrv.bib,References.bib}

\end{document}